# Face Recognition using 3D Facial Shape and Color Map Information: Comparison and Combination

Afzal Godil, Sandy Ressler and Patrick Grother
National Institute of Standards and Technology, Gaithersburg, MD 20899
{godil, sressler, pgrother }@nist.gov

**ABSTRACT**

In this paper, we investigate the use of 3D surface geometry for face recognition and compare it to one based on color map information. The 3D surface and color map data are from the CAESAR anthropometric database. We find that the recognition performance is not very different between 3D surface and color map information using a principal component analysis algorithm. We also discuss the different techniques for the combination of the 3D surface and color map information for multi-modal recognition by using different fusion approaches and show that there is significant improvement in results. The effectiveness of various techniques is compared and evaluated on a dataset with 200 subjects in two different positions.

## 1. INTRODUCTION

Face recognition is of great importance in many applications such as personal identification, employee access to high-security areas, human-machine interfaces, and image retrieval. The main advantage of face recognition as a biometric is its throughput, convenience and non-invasiveness. Most of the research in face recognition has focused on 2D intensity images. However, results from recent studies, FERET [ 4, 5 ] and the FRVT 2002 [ 3 ] clearly show that the performance of these traditional 2D  face recognition approaches are adversely affected by varying lighting conditions and particularly with respect to varying pose. However, the 3D facial surface information is the explicit representation of 3D shape and is invariant under both different lighting and pose conditions. The problem of varying pose can also be corrected relatively easily by rotating the 3D facial surface around the symmetry lines. Hence the next logical step to counter these problems is to use 3D surface information along with the color map of the human face for analysis and comparisons.

In this paper, we investigate a face recognition system based on 3D facial surface information to perform identification and verification, in any facial pose and lighting variations. We then compare the results to one based on color map information and finally study the performance available by fusing the two.

In our study we neglect the effect of facial expression and perform face recognition using 3D facial shape and color map information by using the 3D surface grid and color map information from the CAESAR [ 1, 2 ] anthropometric database. First we use four anthropometric landmark points on the face from the database to properly position and align the face surface and then interpolate the surface information and color map on a regular rectangular grid whose size is proportional to the distance between the landmark points. The grid size is 128 in both directions. We use a cubic interpolation and handle missing values with the nearest neighbor method when there are voids in the original grid. Next we perform principal component analysis (PCA) on the 3D surface and color map information and similarity measures are created using a classifier based on L1 norm. We follow the FRVT methodology and use similarity matrices to compute identification and verification performance scores. We also discuss different techniques for the combination of the 3D surface and color map information for multi-modal recognition. The techniques considered are fusion at the image level and score level. The image level fusion is created by concatenation of the 3D shape and color map information. The score level fusion combines scores using min, max, mean and product rule. The effectiveness of various techniques is compared and evaluated on a dataset with 200 subjects in two different positions, standing and sitting from the CAESAR anthropometric database which was obtained using 3D laser scanning.

This paper is organized as follows. Section 2 describes the previous work in 3D face recognition. In section 3 we discuss 3D capture methods and talk about the CAESAR anthropometric database. In section 4, we discuss the normalization and generation of the facial grid. 3D face recognition is discussed in section 5. In section 6, multi-modal biometric is discussed. Section 7 describes the experiments performed. In section 8, we discuss the Recognition performance. The experimental results are presented in section 9, and conclusions are drawn in section 10.

## 2. PREVIOUS WORK

### 2.1 3D Face Recognition

Face recognition systems based on 3D facial surface information to improve the accuracy and robustness with regard to facial pose and lighting variations have not been addressed thoroughly. Only a few works on the use of 3D data have been reported. Initial studies concentrated on the curvature analysis [3,4,5]. Gordon [7, 8] presented a template based recognition method involving curvature calculation from range data. Lee et. al. [9] proposed a method based on Extended Gaussian Image for matching graph of range images. A method to label different components of human faces for recognition was proposed by Yacoob et. al.[10]. Chua et. al. [11] described a technique based on point signature, a representation for free form surfaces. Beumier et al [12, 13] proposed two 3D different methods based on surface matching and profile matching. Blantz and Venter[15, 16] have used a 3D morphable model to tackle variation of pose and illumination in face recognition, in which the input is a 2D face image. Recently Pan et. al. [14] have used Hausdorff distance for aligning and comparing for 3D recognition. More recently Chang et. al.[17, 18] have used PCA with 3D range data along with 2D image for face recognition.

## 3. 3D CAPTURE

There are three main ways for 3D facial surface capture. The first one is based on structured lighting, in which a pattern is projected on a face and the 3D facial surface is calculated. The second is passive stereo using two cameras to capture a facial image and using a computational matching method, the 3D facial surface is created. Finally the third method is based on the use of laser range-finding systems to capture the 3D facial surface.

The 3D facial surface data quality is not as good as the 2D colored images from a digital camera. The reason is that the 3D data usually have missing data or voids in the concave area of a surface, eyes, nostrils and areas with facial hair. These issues do not happen to a image from a digital camera. The facial surface data available to us from the CAESAR database is also coarse (~4000 points) compared to a 2D image ( 3 to 8 million pixels ) from a digital camera and also compared to other 3D studies [17, 18], where they had around 200,000 points on the facial surface area. The cost of a 3D scanner is also much higher compared to a digital camera for taking 2D images.

The data for our 3D face recognition is from the CAESAR anthropometric database in which 5000 people were scanned using a laser range-finding system. Details about the CAESAR database are discussed in the following subsection.

### 3.1 CAESAR database

The CAESAR (Civilian American and European Surface Anthropometry Resource) project has collected 3D Scans, seventy-three Anthropometry Landmarks, and Traditional Measurements data of 5000 people. The objective of this study was to represent, in three-dimensions, the anthropometric variability of the civilian populations of Europe and North America. The CAESAR project employs both 3-D scanning and traditional tools for body measurements for people ages 18-65. A typical CAESAR body is shown in Figure 1.

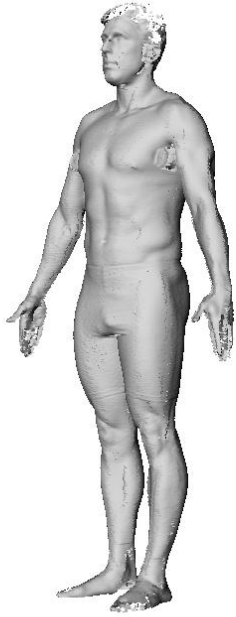
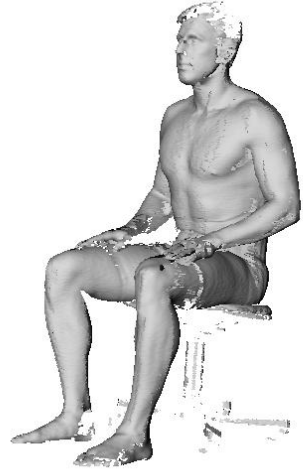

Figure 1a. A typical CAESAR body in standing position    Figure 1b. A typical CAESAR body in sitting position

The seventy-three Anthropometry Landmarks were extracted from the scans. These are point-to-point distances where the points are pre-marked by pasting small stickers on the body and automatically extracted using landmark software. There are around 200,000 points in each surface grid on the body and points are distributed uniformly.

## 4. 3D SURFACE NORMALIZATION

First we cut part of the facial grid from the whole CAESAR body grid using the landmark points 5 and 10 as shown in Figure 2. In Table 1, has the numbers and names of landmark points used in our 3D face recognition study. The new generated facial grid for some of the subjects with two different views is shown in figure 3. Figures 4 a and b, show the histogram of number of grid points in the facial grid. In the case of people standing the minimum number of grid points is 2445 and the mean number is 5729. For the case where people are sitting the minimum number of grid points in the facial surface is 660 and the mean number is 4533. It shows that the grid is very coarse for some of the subjects in the seated pose.

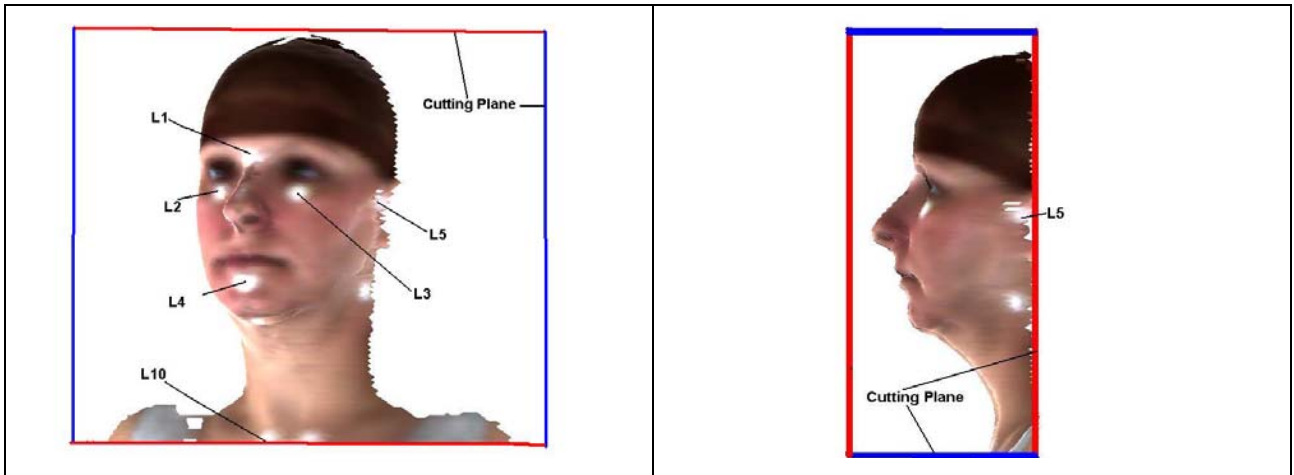

Figure 2. Landmark points 1, 2, 3, 4, 5 and 10. Vertical and horizontal lines are the cutting plane

Table 1. Numbers and names of Landmark point used in our 3D face

| 1   Sellion | 2.  Rt Infraobitale | 3. Lt Infraobitale | 4. Supramenton |
|---|---|---|---|
| 5.  Rt. Tragion | 6.  Rt. Gonion | 7. Lt. Tragion | 8. Lt. Gonion |
| 10. Rt. Clavicale | 12. Lt. Clavicale | | |

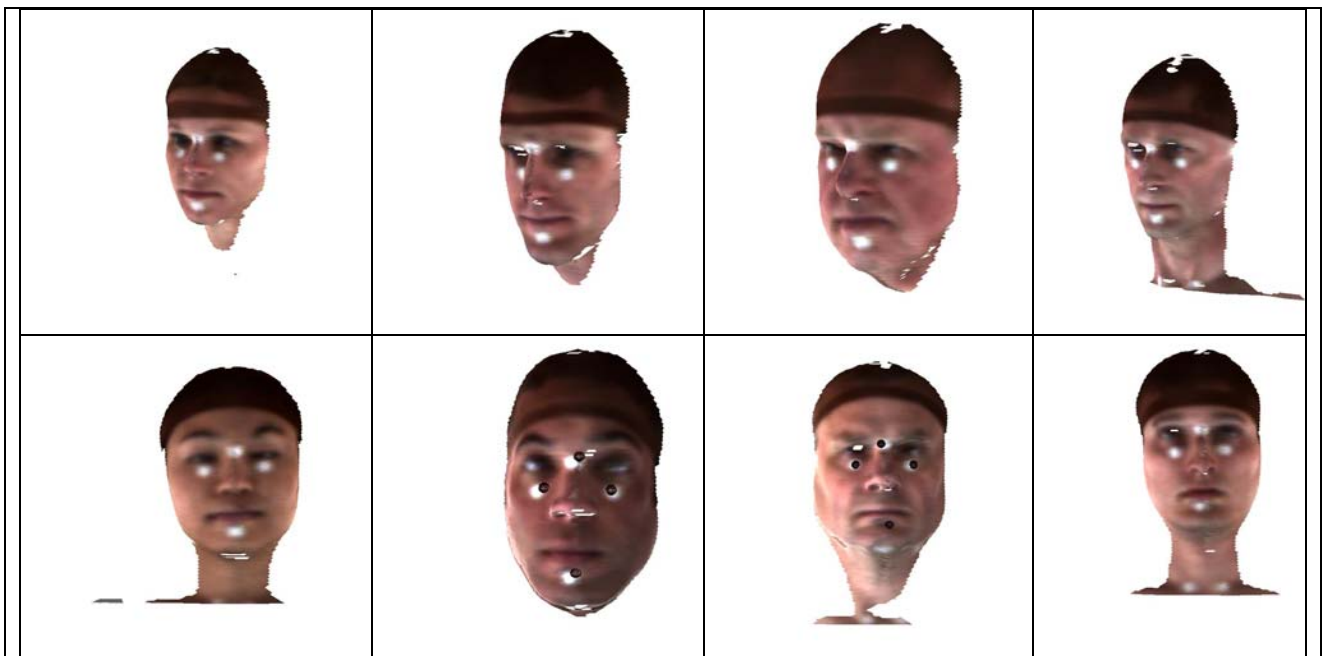

Figure 3. Facial surfaces after the cut from the CAESAR body in two different views.

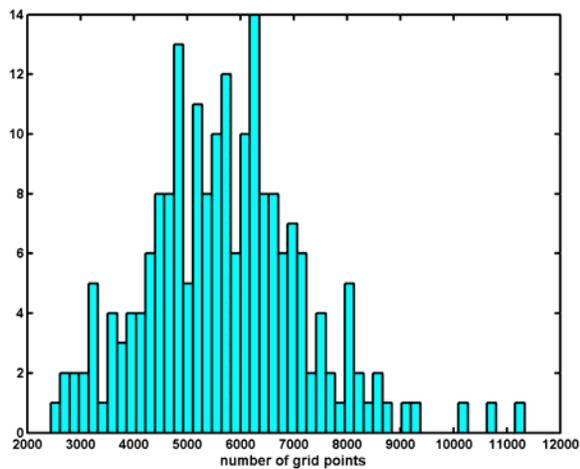

Figure 4a. Histogram of number grid points vs. number of subjects, in the original facial surface (person standing)

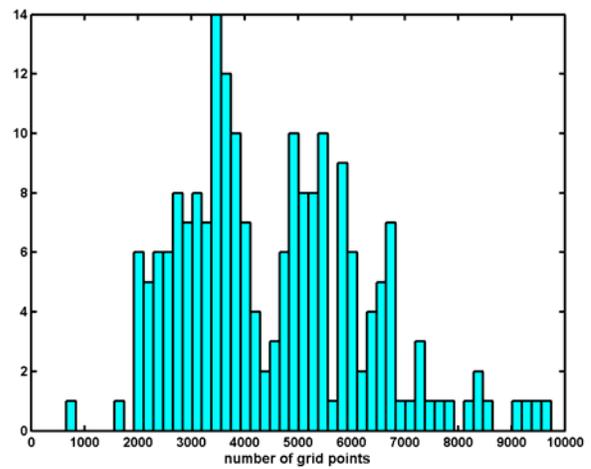

Figure 4b. Histogram of number grid points vs. number of subjects, in the original facial surface (person sitting)

Next, we use four anthropometric landmark points (L1, L2, L3, L4) as shown in figure 2., located on the facial surface, to properly position and align the face surface using an iterative method. There is some error in alignment and position because of error in measurements of the position of these landmark points. Then we interpolate the facial surface information and color map on a regular rectangular grid whose size is proportional to the distance between the landmark points L2, L3, $d=| L3 - L2 |$ and the grid size is 128 in both directions. We use a cubic interpolation and handle missing values with the nearest neighbor method when there are voids in the original facial grid. For some of the subjects there are large voids in the facial surface grids. Figure 5, shows the facial surface and the new rectangular grid.

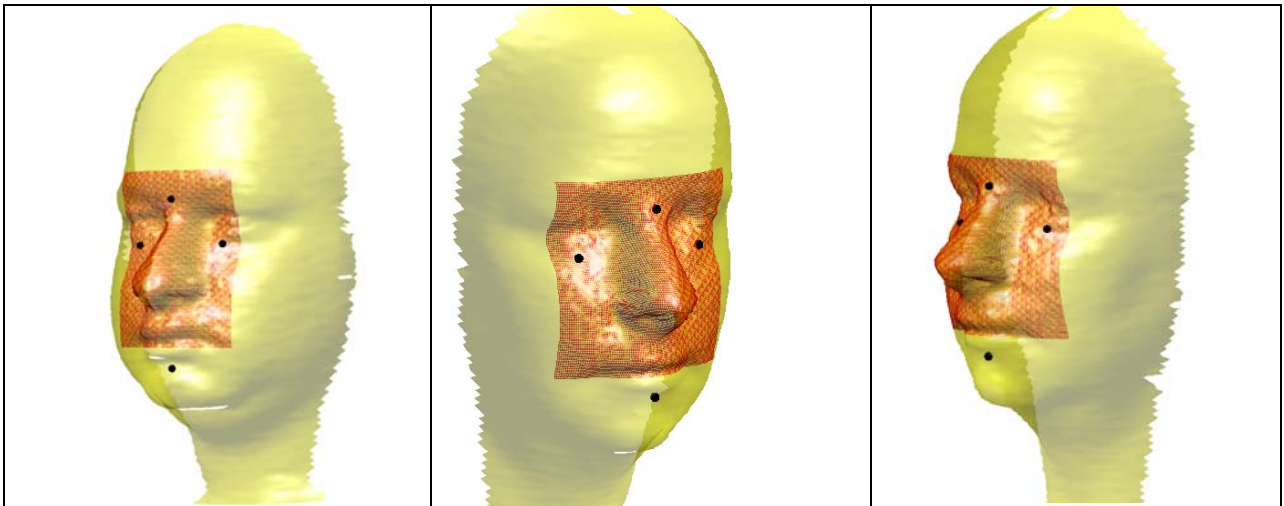

Figure 5. Shows the new facial rectangular grid for three subjects

## 5. 3D FACE RECOGNITION

### 5.1 PCA algorithm

The PCA algorithm is widely employed in traditional face recognition [19]. We will describe it for clarity. The PCA subspace is formed by a scatter matrix formed by a training set of images or from 3D surface information. A set of k training images or 3D surface are used. The PCA recognition method is a nearest neighbor classifier operating in the PCA subspace. Both the training and testing images or 3D facial surface are compared in the PCA subspace.

## 5.2 Distance measure
Two commonly used distances are tested in our study:

The L1 distance $$d(S_i, S_j) = \sum_{i=1}^{k} |S_i - S_j|$$

and the Mahalanobis distance $$d(S_i, S_j) = -\sum_{i=1}^{k} (1/\sqrt{\lambda_i}) S_i S_j$$

Where $\lambda_i$ is the ith eigenvalue corresponding to the ith eigenvector.

## 6. MULTIMODAL BIOMETRIC

The performance of a biometric is affected by the error of the sensor and degrees of freedom provided by the sensor. These problem can be alleviated by the use of multiple sensors, such systems are known as multimodal biometric systems. These are generally more reliable due to the presence of multiple pieces of evidence. There are various types of fusion that are possible when combining multiple sensors: 1) fusion at image level, where features from different sensor are concatenated; 2) fusion at feature level, combines various features; 3) fusion at scoring level, where matching scores are combined and ; 4) fusion at decision level, where accept/reject decisions of multiple systems are consolidated. Early stage fusion is known to outperform later stage fusion.

In our study we have tried two different techniques for the combination of the 3D surface and color map information for multimodal recognition. The techniques considered are fusion at the image level where concatenation of the 3D shape and color map information are formed. We also have explored fusion at the scoring level with such as min, mean, max and product rule.

### 6.1 Score Normalization
An important part is the normalization of scores obtained from different classifiers. Normalization typically involves mapping the scores obtained from multiple frameworks into a common scale and range before combining them [20-25]. This could be viewed as a two step process in which the scores of distribution of each biometric are estimated using statistical methods and then these score distributions are mapped and translated into a common range.

MinMax $\quad S^n = (S - \min(S))/(\max(S) - \min(S))$

Zscore $\quad S^n = (S - mean(S))/std(S)$

Both schemes are linear, and the statistical quantities are usually estimated empirically. The Zscore is usually preferred since the sample extreme values (max and min) are clearly sample dependent and non-robust.

### 6.2 Fusion
We have tried a few of the well-known fusion techniques [20-25], such as the mean, min, max and product rule[ 25].

Mean or Sum Rule $\quad S^j = (S_{3d} + S_{2d})/2$

Min Rule $\quad S^j = \min(S_{3d}, S_{2d})$

Max Rule $\quad S^j = \max(S_{3d}, S_{2d})$

Product Rule $\quad S^j = product(S_{3d}, S_{2d})$

# 7. EXPERIMENTS

In this study we have performed the following experiments (1) testing the hypothesis that CAESAR Anthropometric database can be used for 3D face recognition, in spite of the fact the grid is coarse in the facial area, (2) compare the performance of face recognition system based on 3D surface information and compare it to the one based on color map information using a PCA based method, and (3) test simple fusion based approaches for combining 3D face and color map for improving the performance of face recognition.

In this discussion, the gallery is the group of enrolled biometric signatures and probe set refers to the group of unknown test signatures. For the gallery we use 200 standing subjects from the CAESAR database and 200 sitting subjects for the probe set. The PCA base recognition then simply computes the L1 or Mahalanobis distance between all pair of the i-th gallery and j-th probe signature to a from the similarity matrix based on the 3D shape information and the color map information.

# 8. RECOGNITION PERFORMANCE

We have followed the FRVT 2002 [3, 4, 5] methodology and have simply selected the facial 3D surface and color map information of subjects standing in the gallery and have selected the 3D surface and color map information of subjects sitting in the probe set. Next we perform PCA on the 3D surface and color map information and similarity measure matrices are created using a classifier.

The resulting matrices are used to compute the identification and verification performance scores. The standard measure of verification performance is Receiver Operating Characteristic (ROC). The ROC plot shows the false alarm rate (FAR) on the horizontal axis and the probability of verification on the vertical axis, which is also one minus the false reject rate (1-FRR). FAR is the percentage of imposters wrongly accepted by the security system while FRR is the percentage of valid users rejected by the security system. Hence there is tradeoff between FAR and FRR that depends on security policy and throughput requirements.

The measure of identification performance is the "rank order statistic" called the Cumulative Match Characteristic (CMC). The rank order statistics indicate the probability that the gallery subject will be among the top r matches to a probe. This probability depends upon both, gallery size G and r.

# 9. RESULTS

The results for the 3D, 2D face recognition and fusion are presented below.

## 9.1. Results for 2D and 3D

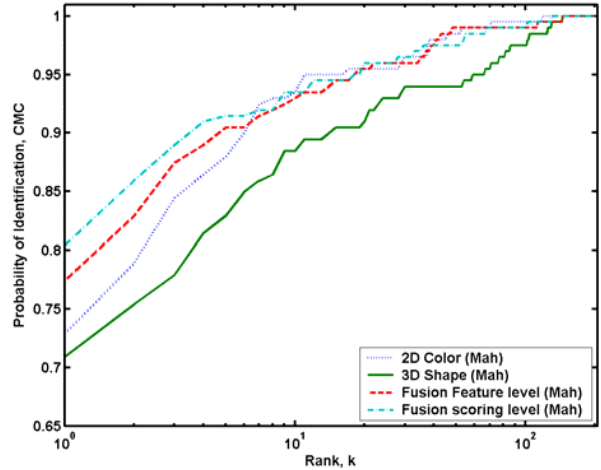 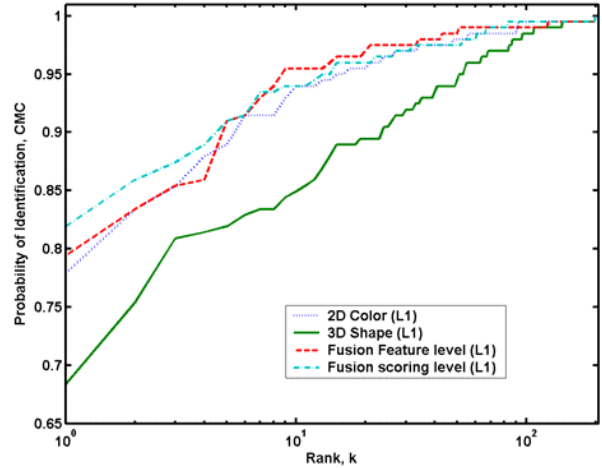

Figure 6. The Cumulative Match Characteristic curves for color map, 3D shape and fusion at image level and scoring level with Mahalanobis based classifier

Figure 7. The Cumulative Match Characteristic curves for color map, 3D shape and fusion at image level and scoring level with L1 based classifier

The evaluation performed on the data based on two metrics, Identification Performance based on Cumulative Match Characteristic is shown below. The plot shows the fractions of probe signatures whose gallery match was within the given ranks.

In Figure 6, we see the CMC for signature based on 2D color map, 3D facial shape, fusion based on image level and fusion based on score level is compared with a Mahalanobis based classifier. The plot shows the CMC at rank one for 2D map is .728, for 3D shape it is .708 , for fusion at image level is .7738 and CMC (1) for fusion at scoring level is .81. For the case of fusion at scoring level we used Z-score normalization and fusion based on mean of the 2D and 3D similarity matrix.

In Figure 7, we see the CMC for signature based on 2D color map, 3D facial shape, fusion based on image level and fusion based on score level is compared with a L1 distance based classifier. The plot shows the CMC at rank one for 2D map is .778, for 3D shape it is .683, for fusion at image level is .794 and CMC (1) for fusion at scoring level is .82. For the case of fusion at scoring level we used Z-score normalization and fusion based on mean of the 2D and 3D similarity matrix

The results for figures 6, 7 clearly show that we get reasonable results from the coarse grid for both 2D color map and 3D shape signature. The results from 2D color map are better that that from the 3D facial shape signature. The fusion from the two biometric also show better performance compared to the individual biometric. The results also show that fusion at scoring level performs much better then fusion at image level.

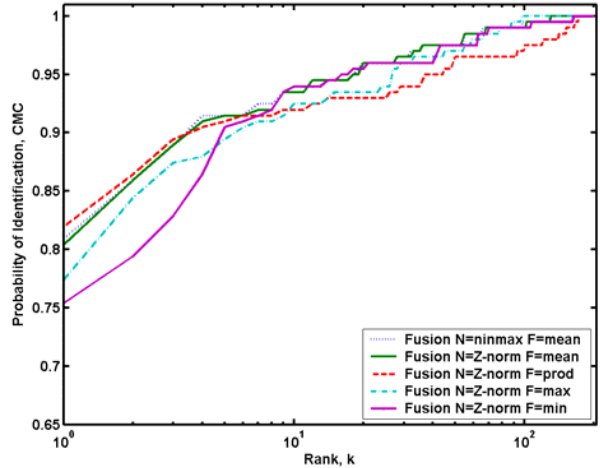 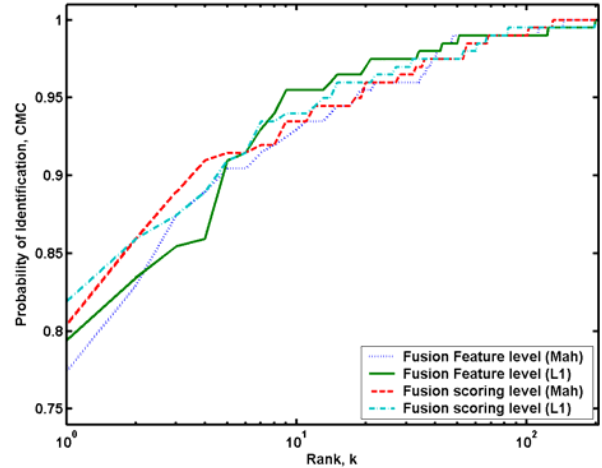

Figure 8. The CMC curves with different fusion rule and normalization.

Figure 9. The Cumulative Mach Characteristic curves for fusion with image and scoring level with L1 and Mahalanobis based classifier

The CMC for different types of fusion and normalization is shown in Figure 8. The plot shows that the MinMax normalization performs better that to Z-score normalization for mean rule fusion. For the rest of the fusion rule test we used the Z-score normalization and best performance is with the product rule fusion, second is the mean rule, third and fourth are max and min rule. In spite of the fact that the performance with MinMax normalization is better than Z-score normalization, for our study, we mainly still mainly use Z-score normalization because it is less effected by round off error.

In Figure 9 the Cumulative Mach Characteristic curves for fusion at image level and scoring level with L1 and Mahalanobis based classifier are compared for Z-score normalization for fusion at scoring level.

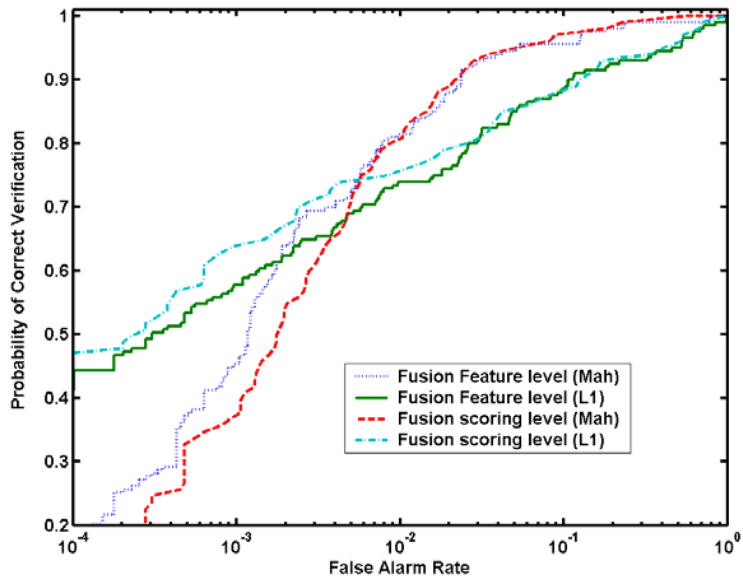

Figure 10. The ROC graph for fusion with image and scoring level with L1 and Mahalanobis based classifier

Next we show verification performance of the biometric using the Receiver Operating Characteristics (ROC) as shown in Figure 10. The ROC graph shows the true accept rate (legitimate access) vs. the false accept rate (erroneous admission) and shows ROC curves for fusion at image level and scoring level with L1 and Mahalanobis based classifier for Z-score normalization for fusion at scoring level. The plot show that 80% of persons are correctly verified while falsely accepting 1%. The plot also show that performance with L1 based classifier is much better that one based on Mahalanobis based classifier at lower FAR such as at .1%. However the performance Mahalanobis based classifier is much higher that with L1 classifier at higher FAR value such as 2%.

## 10. CONCLUSION

In this study we have shown that it is possible to use facial surface and color map information from the CAESAR database for face recognition. The performance based on color map signature is slightly better than that based on 3D facial surface signature. We also have shown that fusion of the two biometrics gives higher performance than individual biometrics. The results show that the performance of fusion at score level is higher than of fusion at image level. We suspect that this due to a scaling problem in the 3D and color map data.

The quality of 3D surface data is not as good as the 2D colored images from a digital camera. The 3D data usually have missing data or voids in the concave area of a surface, eyes, nostrils and areas of facial hair that does not happen with a digital camera. The facial surface data available to us from the CAESAR database is also coarse (~4000 points) compared to a 2D image ( 3 to 8 million pixels ) from a digital camera and also compared to other 3D studies [17,18], where they had around 200,000 points on the facial surface area. The cost of a 3D scanner is also much higher than a digital camera for 2D images.

Chang et. al.[17, 18] have recently obtained much higher performance than our results, because of better facial grid with more grid points, and possibly because of a different 3D face normalization method.

In the future, we will study other recognition methods for 3D face recognition and also try other methods of fusion to improve the performance of the system. We will also use more subjects in our experiments as we have 5000 bodies available to us.

## ACKNOWLEDGEMENT

We would like to thank Dr. Kathleen Robinette of Wright - Patterson Air Force Base, Dayton, USA for providing us the CAESAR Anthropometry Database. We would also like to thank Qiming Wang of NIST for her help during this project.

## REFERENCES

1. CAESAR web site: http://www.sae.org/technicalcommittees/caesumm.htm
2. Civilian American and European Surface Anthropometry Resource (CAESAR) web site: http://www.hec.afrl.af.mil/cardlab/CAESAR/index.html
3. P. Phillips, P. Grother, R. Micheals, D. Blackburn, E. Tabassi and M. Bone. Face Recognition Vendor Test 2002. In NIST Technical Report, NIST IR 6965, March 2003.
4. S. Rizvi, P. Phillips and H. Moon. The FERET Verification Testing Protocol for Face. In NIST Technical Report, NIST IR 6281, October 1998.
5. P. Phillips, H. Moon, S. Rizvi, and P. Rauss. The FERET Evaluation Methodology for Face-Recognition Algorithms. In IEEE Trans. Pattern Analysis and Machine Intelligence, 22:1090-1103, 2000.
6. G. Gordon. Face Recognition from Frontal and Profile Views. In Proceedings of the International Workshop on Face and Gesture Recognition , (Zurich, Switzerland), pp.47--52, June 1995.
7. G. Gordon. Face Recognition Based on Depth and Curvature Features. In Proceedings of the IEEE Computer Society Conference on Computer Vision and Pattern Recognition, (Champaign, Illinois), pp.108-110, June 1992.
8. G. Gordon, L. Vincent. Application of Morphology to Feature Extraction for Face Recognition. In Proc. of SPIE, Nonlinear Image Processing, San Jose, Feb. 1992. Vol. 1658.

9. J. C. Lee and E. Milios. Matching rang Images of human faces. In proc. IEEE International Conference on Computer Vision, Page 722-726, 1990.
10. Y. Yacoob and L. S. Davis. Labeling of human faces components from range data. CVGIP: Image Understanding, 60(2): 168-178, September 1994.
11. C.S. Chau , F. Han and Y. K. Ho. 3D face recognition using point signature. In Proc. IEEE International Conference on Automatic Face and Gesture Recognition, Pages 233-238, March 2000.
12. C. Beumier, M. Acheroy. Automatic Face Verification from 3D and Grey Level Clues In RECPAD2000, 11th Portuguese conference on pattern recognition, Porto, Portugal, May 11-12, 2000, pp 95-101.
13. C. Beumier, M. Acheroy. Face Verification from 3D and grey-level clues. In Pattern Recognition Letters, Vol. 22, 2001, pp 1321-1329.
14. G. Pan, Y. Wu, Z. Wu. Investigating Profile extraction from range data for 3D recognition. In IEEE International Conference on Multimedia & Expo (ICME), july6-9,2004.
15. V. Blanz, T. Vetter. Face Recognition Based on Fitting a 3D Morphable Model. In IEEE Transactions on Pattern Analysis and Machine Intelligence 25(9): 1063-1074 (2003)
16. V. Blanz, T. Vetter. A Morphable Model for the Synthesis of 3D Faces. SIGGRAPH 1999. 187-194.
17. K. Chang, K. Bowyer, P. Flynn. Face Recognition Using 2D and 3D Facial Data. In IEEE International Workshop on Analysis and Modeling of Faces and Gestures. Nice, France. Oct. 2003.
18. K. Chang, K. W. Bowyer, and P. J. Flynn. Face recognition using 2D and 3D facial data, Workshop on Multimodal User Authentication (MMUA), December 11-12 2003
19. L. Sirovich and M. Kirby. A low-dimensional procedure for the characterization of human faces. Journal of the Optical Society of America, 4(3):519--524, 1987
20. M. Indovina, U. Uludag, R. Snelick, A. Mink and A. Jain. Multimodal Biometric Authentication Methods: A COTS Approach, Proc. MMUA 2003, Workshop on Multimodal User Authentication, pp. 99-106, Santa Barbara, CA, December 11-12, 2003.
21. A. Ross and A.K. Jain. Information Fusion in Biometrics. Proc. of AVBPA, Halmstad. Sweden. pp. 354-359, June 2001.
22. B. Achermann and H. Bunke. Combination of classifiers on the decision level for face recognition. Technical Report IAM-96-002, Insitut fur Informatik und angewandte Mathematik, Universitat Bern, Bern, 1996.
23. B. Brunelli and D. Flavigna. Personal identification using multiple cues. IEEE Transactions on Pattern Analysis and Machine Intelligence, 17(10):955–966, 1995.
24. L. Hong and A. Jain. Multimodal biometrics. In A. Jain, R. Bolle, and S. Pankanti, editors, Biometrics: Personal Identification in Networked Society. Kluwer, 1999.
25. J. Kittler, M. Hatef, R. Duin, and J. Matas. On Combining Classifiers. IEEE Transactions on Pattern Analysis and Machine Intelligence, 20(3):226–239, March 1998.